 \documentclass[final,5p,times,twocolumn,authoryear,table]{elsarticle}

\usepackage{amssymb}

\journal{Computers and Electronics in Agriculture}

\usepackage{tikz}
\usetikzlibrary{arrows.meta}
\usepackage{pgfplots}
\pgfplotsset{compat=newest}
\pgfplotsset{plot coordinates/math parser=false}
\newlength\figureheight
\newlength\figurewidth

\usepackage{todonotes}
\usepackage{booktabs}

\usepackage{amsmath}

\begin{document}

\begin{frontmatter}



\title{Apple scab detection in orchards using deep learning on colour and multispectral images}


\affiliation[inst1]{organization={Department of Informatics at Faculty of Business and Economics, Mendel University in Brno},
            addressline={Zemedelska 1665/1}, 
            city={Brno},
            postcode={613\,00},
            country={Czech Republic}}

\affiliation[inst2]{organization={Agro-Food Robotics, Wageningen University \& Research, },
            addressline={Droevendaalsesteeg 1},
            postcode={6708 PB},
            city={Wageningen},
            country={the Netherlands}}

\affiliation[inst3]{organization={Farm Technology Group, Wageningen University \& Research, },
            addressline={Droevendaalsesteeg 1},
            postcode={6708 PB},
            city={Wageningen},
            country={the Netherlands}}
            
\author[inst1]{Robert Rou\v{s}}
\author[inst2]{Joseph Peller}
\author[inst2]{Gerrit Polder}
\author[inst2]{Selwin Hageraats}
\author[inst3]{Thijs Ruigrok}
\author[inst2]{Pieter M. Blok}

\begin{abstract}
Apple scab is a fungal disease caused by \textit{Venturia inaequalis}. Disease is of particular concern for growers, as it causes significant damage to fruit and leaves, leading to loss of fruit and yield. This article examines the ability of deep learning and hyperspectral imaging to accurately identify an apple symptom infection in apple trees. In total, 168 image scenes were collected using conventional RGB and Visible to Near-infrared (VIS-NIR) spectral imaging (8 channels) in infected orchards. Spectral data were preprocessed with an Artificial Neural Network (ANN) trained in segmentation to detect scab pixels based on spectral information. Linear Discriminant Analysis (LDA) was used to find the most discriminating channels in spectral data based on the healthy leaf and scab infested leaf spectra. Five combinations of false-colour images were created from the spectral data and the segmentation net results. The images were trained and evaluated with a modified version of the YOLOv5 network. Despite the promising results of deep learning using RGB images (P=0.8, mAP@50=0.73), the detection of apple scab in apple trees using multispectral imaging proved to be a difficult task. The high-light environment of the open field made it difficult to collect a balanced spectrum from the multispectral camera, since the infrared channel and the visible channels needed to be constantly balanced so that they did not overexpose in the images.

\end{abstract}




\end{frontmatter}




\section{Introduction}\label{sec:introduction}


According to FAOSTAT \citep{fao2022} worldwide production of apples in the last 10 years averages more than 80 million tons annually. In terms of production, apple ranks third. Therefore, apple is a significant commercial fruit crop, and any research and development in the apple growing industry has an economic potential worthy of attention.

Apple scab is a fungal disease caused by \textit{Venturia inaequalis} that affects apple trees and fruits. The disease is of particular concern for growers as it causes significant damage to fruit and leaves, leading to reduced yield and marketability. Traditionally, the evaluation of diseases and stresses is based on visual inspection and destructive diagnostic methods. However, these methods are time-consuming and labour intensive and do not allow for early detection of the disease. Automatic detection could alleviate this problem, while simultaneously reducing the application of plant protection agents.

In the last decade, much work has been done on the spectral imaging of diseases in agriculture. Spectral imaging allows for direct detection of pests in the plant, as well as changes in the physiology of the plant as it is infected. Studies have shown that spectroscopy can be used to detect mildew formation in leaves as the mycelium penetrates the leaf. Therefore, it is possible to use spectral imaging to detect such diseases in wine grapes \citep{Knauer2017} or wheat \citep{Zhang2016}. Spectral imaging can also be used to detect chlorosis caused by necrotrophic diseases such as \textit{Alternaria} \citep{Vijver2020} or scab in short-wave infrared (SWIR) spectra \citep{Gorretta2019} or VIS-NIR \citep{Solovchenko2021}. Stress-induced biophysical and biochemical modifications will directly affect leaf reflectance obtained by spectral imaging. However, simply detecting these changes in the spectrum of the plant is often not enough, as these symptoms can be caused by multiple factors. Therefore, it is necessary to combine these techniques with machine learning and deep learning.

Deep learning has been extensively researched and developed in the past decade. In particular, convolutional neural networks (CNNs) are being used increasingly for disease and pest detection applications \citep{kamilaris2018}. The advantage of CNNs over traditional machine learning methods is their ability to automatically discriminate and learn features that would be difficult to be hand-crafted by humans \citep{LeCun2015}. Succesful applications of CNNs for plant disease detection include fusarium head blight detection in wheat \citep{Qiu2019}, blackleg detection in potato \citep{AFONSO2019}, anthracnose detection in olives \citep{Fazari2021} and the detection of alternaria leaf blotch and rust disease in apple leaves \citep{Bi2020}. 

Many recent articles use publicly available image data, such as the PlantVillage dataset \citep{Hughes2015} to train CNNs on RGB images to detect various diseases. \citet{Mohanty2016} or \citet{Geetharamani2019} use these data to compare different types of neural network for disease detection in general, \citet{Liu2017} uses them to detect apple leaf diseases. Their results show that this approach is valid for this kind of task with an overall accuracy greater than 95\%, but real-world applications can bring about more challenging situations. Such as changing weather conditions and thus not-stable lighting conditions. This problem can be mitigated by using custom imaging systems with top-view camera perspective that also allow the camera to be shielded by a box (with artificial light) and are moved over imaged plants. However, this is applicable only for crops such as potato plants \citep{Vijver2020, Polder2019}. Spectral imaging under field conditions for crops such as apple trees can be performed from the top with an unmanned aerial vehicle (UAV) or by using side-view imaging as proposed by \citet{Karpyshev2021} or \citet{Nguyen2021}. Field conditions with natural light that also require the use of white reference \citep{Gutierrez2018, Paulus2020} are challenging due to the setting of the exposition time. The application of object detection also depends on object occlusion. If a real-time application is needed, for example, implementation into mobile agricultural machinery, the inference speed is also an important factor, and the resulting application is usually a trade-off between speed and accuracy.

Relying only on "object detection" methods from conventional RGB cameras can have disadvantages in terms of misinterpreting naturally occurring shapes in healthy plants as a disease.

Attempts have been made to incorporate spectral imaging and deep learning. More than 30 publications have dealt with the application of deep learning in spectral imaging in agriculture in recent years \citep{Wang2021}. However, the large number of image layers in a spectral data array makes direct network training difficult.

A reliable, accurate, and nondestructive measure is essential to quickly detect the incidence of diseases in crops to allow for timely intervention to prevent disease from spreading across the field. This research article investigates the use of spectral images in neural networks to detect apple scab in orchards and assesses their potential for effective disease management. Specifically, the article examines the ability of deep learning and hyperspectral imaging to accurately identify apple scab on apple trees, as well as the potential of these technologies to improve orchard management strategies. The implications of the findings for orchard management are also discussed.


\section{Materials and Methods}
\subsection{Imaging system and acquisition}
\begin{figure}[htbp]
    \centering
    \includegraphics{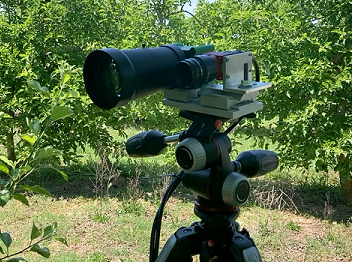}
    \caption{Imaging camera setup from the field. The Multispectral camera and RGB camera had overlapping fields of view and were triggered together using a software trigger.}
    \label{fig:imagingSetup}
\end{figure}

An imaging platform was deployed in the field with several cameras. A camera plate was designed to fit on a standard tripod and was controlled via a laptop computer. A RGB camera, an IDS 10 Mpixel NXT camera, and a SILIOS CMS4-V eight-band spectral camera were attached to the plate. The SILIOS acquires images in nine channels, eight for each specific wavelength (545, 579, 622, 658, 701, 737, 779, and 816\,nm) and one PAN (panchromatic) channel measuring the light intensity across all eight channels. The resolution of the spectral images is 682 by 682 pixels. Both cameras were equipped with the same Tamron 8 mm adjustable lens and recorded an overlapping field of view. The imaging system is shown in Figure \ref{fig:imagingSetup}.

\subsection{Samples and experiment location}
Images were collected in apple orchards outside of Épila, Spain, in May 2022. Scouting was performed with the assistance of crop experts who helped identify areas of scab for imaging. Once an area was identified, a set of images was collected so that each RGB image had a corresponding spectral image. To ensure that the apple scab could be easily identified, an additional image with the expert pointing to the visible scab was also collected. In total, 168 image scenes were acquired. Each of these scenes contained one or more areas of apple scab. The 168 image scenes cover a total of 325 areas of apple scab.

\subsection{Spectral data (pre)processing}
In this research, the purpose of using spectral imaging data was to enhance the contrast between healthy and symptomatic leaf tissue, compared to the contrast captured by a regular RGB camera. Two tailored machine-learning based image processing procedures were developed to convert the eight-channel spectral images into images of reduced dimensionality that specifically highlight areas that are spectrally similar to scab symptoms.  

Prior to any dimensionality reduction procedures, all channels in the spectral images were first normalised using linear coefficients obtained from the PAN channel intensity of a white colour checker board. The coefficients were calculated from Equation \ref{eq:normalisation} for each channel for every dataset where the calibration image was taken.

\begin{equation}\label{eq:normalisation}
\text{\textit{norm.~coefficient}}_{channel} = \frac{channel~intensity}{PAN~intensity}
\end{equation}

Then, in order to provide the two machine learning-based image processing methods with enough information on the spectral differences between healthy and symptomatic leaf tissue, ten normalised eight-channel spectral images were annotated by drawing masks over the visible scab symptoms. This automatically provided class labels for more than four million pixels and the associated eight-channel spectra. 

In the first approach, a linear discriminant analysis (LDA) model was trained on all available labelled spectra. The LDA model weights then provide an indication for the discriminative power of each of the eight spectral channels. By selecting the three channels with the highest weights and combining them into false-colour RGB images, a dataset was generated that could be used for subsequent deep learning-based symptom detection. 

\begin{figure}[htbp]
    \centering
    \includegraphics[]{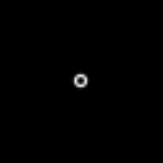}
    \includegraphics[]{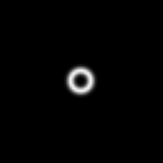}
    \includegraphics[]{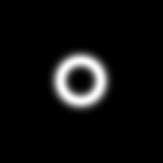}
    \includegraphics[]{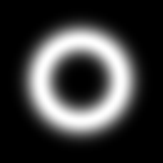}
    \caption{The four kernels used for convolution of the spectral images, generated using Eq. \ref{eq:donut}. From left to right, the parameters used were: $\alpha$ = 50, $\beta$ = 4, 8, 16, 32, and $\sigma$ = 0.781, 1.56, 3.13, 6.25.}
    \label{fig:kernels}
\end{figure}

In the second approach, a simple classification artificial neural network (ANN) was trained on the labelled spectra to discriminate between pixels belonging to symptomatic leaf tissue and pixels belonging to anything else (healthy leaf tissue, branches, soil, etc.). The neural network consisted of four layers: one input layer (8 nodes), two hidden layers (16 \& 8 nodes), and one output layer (2 nodes). Instead of using only the eight output channels of the camera, a series of four spatial convolutions were performed on each of the eight channel images to extend the feature vector to length 40. The four kernels--described by Equation \ref{eq:donut}, and shown in Figure \ref{fig:kernels}---are meant to include some contextual information at different scales, while also providing spectral information at a higher signal-to-noise ratio in the direct vicinity of the central pixel. 

\begin{equation}\label{eq:donut}
value(x,y) = e^{\frac{-1}{\sigma^2}\cdot \sqrt{(x-\alpha)^2 + (y-\alpha)^2} -\beta^2}
\end{equation} 

The four kernels used in this study were calculated with parameters: $\alpha$ = 50, $\beta$ = 4, 8, 16, 32, and $\sigma$ = 0.781, 1.56, 3.13, 6.25, where $\sigma$ is the radial standard deviation, $\alpha$ is half the dimension of the entire kernel matrix, and $\beta$ is proportional to the radius of each kernel's doughnut-shaped intensity profile. 

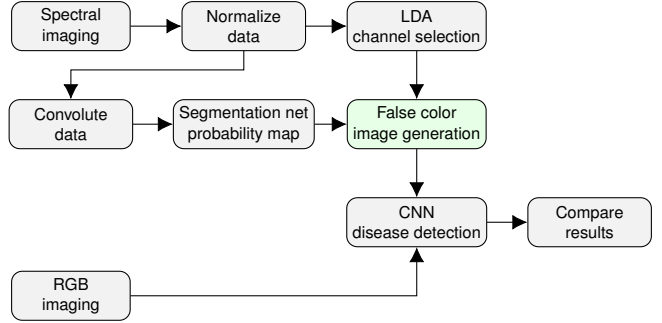
\begin{figure}[htbp]
    \centering
\tikzset{%
  >={Latex[width=2mm,length=2mm]},
            base/.style = {rectangle, rounded corners, draw=black,
                           minimum width=2.4cm, minimum height=1cm,
                           text centered, font=\sffamily},
        activityStarts/.style = {base, fill=gray!10},
        todo/.style = {base, fill=orange!40},
        activityRuns/.style = {base, fill=green!10},
        process/.style = {base, minimum width=2.5cm, fill=gray!10,
                           },
}

\begin{tikzpicture}[node distance=1.5cm,
    every node/.style={fill=white, font=\sffamily}, align=center, scale=0.65, transform shape]
  \node (start)         [activityStarts]                                {Spectral\\imaging};
  \node (Normalize)     [process, right of=start, xshift=2cm]           {Normalize\\data};
  \node (LDA)           [process, right of=Normalize, xshift=2cm]       {LDA\\channel selection};
  \node (pseudoCreate)  [activityRuns, below of=LDA, yshift=-0.5cm]     {False color\\image generation};
  \node (SegNet)        [process, below of=Normalize, yshift=-0.5cm]    {Segmentation net\\probability map};
  \node (Convolute)       [process, below of=start, yshift=-0.5cm] {Convolute\\data};
  \node (CNN)           [process, below of=pseudoCreate, yshift=-0.5cm] {CNN\\disease detection};
  \node (Compare)       [process, right of=CNN, xshift=2cm]             {Compare\\results};
  
  \node (RGB1)          [activityStarts, below of=start, yshift=-4.0cm] {RGB\\imaging};

  \draw[->]             (start) -- (Normalize);
  \draw[->]             (Normalize) -- (LDA);
  \draw[->] (Normalize) |- ([shift={(0mm, 6mm)}] Convolute.north) -- ([xshift= 0mm] Convolute.north);
  \draw[->]             (Convolute) -- (SegNet);
  \draw[->]             (LDA) -- (pseudoCreate);
  \draw[->]             (SegNet) -- (pseudoCreate);
  \draw[->]             (pseudoCreate) -- (CNN);
  \draw[->]             (CNN)  -- (Compare);
  
  \draw[->]             (RGB1)  -| (CNN);

  \end{tikzpicture}
    \caption{Data processing pipeline from image acquisition to object detection}
    \label{fig:dataProcPipe}
\end{figure}

The ANN classification model was trained on a random subset of the 4+ million feature vectors, after which the resulting weights were used to do per-pixel inference on the remaining images. The outputs of the final layer's softmax function for all pixels were then compiled into scab-specific probability maps, which were used to train subsequent deep-learning based symptom detection models. The image data processing pipeline is shown in Figure \ref{fig:dataProcPipe}.

\subsection{False-colour Image Sets}
To compare the efficacy of the neural network using multispectral data, five image sets were created. Each image set was composed of false-colour images comprised of the bands from the multispectral camera or from the mask created from the segmentation network. The priority of the selected bands was based on the strength of each wavelength using the LDA weighting vector. It was used to select image channels in image sets. Different image sets that were compared are shown in Table \ref{tab:ListIma}.

\begin{table*}[htbp]
\caption{list of image sets used for object detection neural network training.}
\begin{center}
    \begin{tabular}{@{}llc@{}}
    \toprule
    Image set & Composition & Total number of channels \\ \midrule
    RGB & RGB image & 3 \\
    SegN & Segmentation Net Mask & 1 \\
    MS7,3,1 & MS bands (779\,nm, 622\,nm, 545\,nm) & 3 \\
    MS7,3+SegN & MS bands (779\,nm, 622\,nm, Segmentation Net Mask) & 3 \\
    \shortstack[l]{MS7,3,1+5,2,6\\~} & \shortstack[l]{Two MS images (779\,nm, 622\,nm, 545\,nm),\\  (701\,nm, 579\,nm, 737\,nm)} & 6 \\
    \shortstack[l]{MS7,3,1+5,2,6+8,4,0\\~} & \shortstack[l]{Three MS images (779\,nm, 622\,nm, 545\,nm), \\(701\,nm, 579\,nm, 737\,), (658\,nm, 816\,nm, grayscale)} & 9 \\
    \hline
    \end{tabular}
\end{center}    
    \label{tab:ListIma}
\end{table*}

\subsection{Deep learning}
The 168 images were divided into training, validation and test set. The training set consisted of 118 images (70\%), and these images were used to optimise the neural network weights during training. The validation set consisted of 25 images (15\%) and these images were used during training to check whether the trained model was overfitting. The test set consisted of 25 images (15\%) and these images were independent of the training process and therefore suitable for evaluation. 

The images were trained and evaluated with a modified version of the You Only Look Once (YOLO) network (version 5) \citep{yolov5}. Our modification involved the customisation of the standard data loader of the YOLOv5 network so that images with more than three image channels could be processed. Our modified data loader stacked the tensors of the input layer in sets of three channels, allowing us to also analyse six-channel and nine-channel images. The input layer of the detection model was changed accordingly to create an input layer that could process three, six, or nine input channels, respectively. The transfer-learnt weights of the first layer were also stacked accordingly from the first three channels of a YOLOv5 network that was pre-trained in the Microsoft Common Objects in Context (COCO) dataset \citep{Lin2014}. We used YOLOv5x as our network architecture. This architecture was considered to provide the highest accuracy according to the results of \citet{yolov5}. 

During training, two types of data augmentation were applied: geometric transformations (rotation, translation, scaling, and horizontal and vertical flips) and image assemblage (merging multiple images into one composite image). The YOLOv5x network was trained with a batch size of 8 images and the number of training epochs was 100. The software of Jocher \citep{yolov5} automatically saved the network weights with the best performance in the validation set. The best performing weights were then used for independent testing. This testing was performed with a threshold of 0.1 on the network confidence level and a threshold of 0.2 on the non-maximum suppression (NMS).  This NMS threshold allowed for marginal overlapping bounding boxes, which was deemed appropriate given our image annotations.   

\section{Results and discussion}

\subsection{Segmentation network results}
\begin{figure}[htbp]
    \centering
%
%
\begin{tikzpicture}

\begin{axis}[%
width=7.2cm,
height=5.1cm,
at={(0.333in,0.303in)},
scale only axis,
xmin=540,
xmax=820,
xtick={545, 579, 622, 658, 701, 737, 779, 816},
xlabel style={font=\color{white!15!black}},
xlabel={wavelength [nm]},
ymin=0.1,
ymax=1,
ylabel style={font=\color{white!15!black}},
ylabel={Normalized reflectance},
axis background/.style={fill=white},
title style={font=\bfseries},
axis x line*=bottom,
axis y line*=left,
xmajorgrids,
ymajorgrids,
legend style={at={(0.03,0.97)}, anchor=north west, legend cell align=left, align=left, draw=white!15!black}
]

\addplot[area legend, draw=none, fill=blue, fill opacity=0.2]
table[row sep=crcr] {%
x	y\\
545	0.141801175050094\\
579	0.157974785065961\\
622	0.169511550526569\\
658	0.177747168404482\\
701	0.185613108243024\\
737	0.198039385930972\\
779	0.214569294568038\\
816	0.209176733695083\\
816	0.70981800864902\\
779	0.732540444422245\\
737	0.700666787143725\\
701	0.578505094244516\\
658	0.509519688673441\\
622	0.496026735279884\\
579	0.474714505333089\\
545	0.457821020420955\\
}--cycle;
\addlegendentry{2 std. dev. $\sigma$}

\addplot [color=white!55!blue, forget plot]
  table[row sep=crcr]{%
545	0.141801175050091\\
579	0.157974785066017\\
622	0.169511550526522\\
658	0.177747168404494\\
701	0.185613108243047\\
737	0.198039385930997\\
779	0.214569294568037\\
816	0.209176733695131\\
};
\addplot [color=white!55!blue, forget plot]
  table[row sep=crcr]{%
545	0.457821020420965\\
622	0.496026735279884\\
658	0.509519688673436\\
701	0.57850509424452\\
737	0.700666787143746\\
779	0.732540444422284\\
816	0.709818008648995\\
};
\addplot [color=blue]
  table[row sep=crcr]{%
545	0.299811097735528\\
579	0.316344645199479\\
622	0.332769142903203\\
658	0.343633428538965\\
701	0.382059101243726\\
737	0.449353086537371\\
779	0.47355486949516\\
816	0.459497371172006\\
};
\addlegendentry{Scab}

\addplot[area legend, draw=none, fill=black, fill opacity=0.2]
table[row sep=crcr] {%
x	y\\
545	0.19321542609238\\
579	0.210394747712112\\
622	0.225718913130852\\
658	0.23477016335149\\
701	0.272537751268131\\
737	0.363700173881785\\
779	0.399010402354098\\
816	0.375771723791977\\
816	0.914928760522262\\
779	0.962247330349749\\
737	0.914875781144891\\
701	0.706224745517086\\
658	0.62159987128075\\
622	0.603057257309427\\
579	0.577229143197011\\
545	0.554356130383077\\
}--cycle;
\addlegendentry{2 std. dev. $\sigma$}

\addplot [color=white!55!black, forget plot]
  table[row sep=crcr]{%
545	0.193215426092365\\
579	0.210394747712144\\
622	0.225718913130891\\
658	0.234770163351527\\
701	0.272537751268146\\
737	0.363700173881739\\
779	0.399010402354065\\
816	0.375771723792013\\
};
\addplot [color=white!55!black, forget plot]
  table[row sep=crcr]{%
545	0.554356130383098\\
579	0.577229143196973\\
622	0.603057257309388\\
658	0.62159987128075\\
701	0.706224745517034\\
737	0.914875781144929\\
779	0.962247330349783\\
816	0.91492876052223\\
};
\addplot [color=black]
  table[row sep=crcr]{%
545	0.373785778237675\\
579	0.393811945454559\\
622	0.414388085220139\\
658	0.428185017316082\\
701	0.489381248392647\\
737	0.639287977513391\\
779	0.680628866351867\\
816	0.645350242157065\\
};
\addlegendentry{Leaves}

\end{axis}

\begin{axis}[%
width=2.835in,
height=2.362in,
at={(0in,0in)},
scale only axis,
xmin=0,
xmax=1,
ymin=0,
ymax=1,
axis line style={draw=none},
ticks=none,
axis x line*=bottom,
axis y line*=left
]
\end{axis}
\end{tikzpicture}%
    \caption{Mean spectra of healthy leaf and scab pixels randomly sampled from 170\,000\,pixels in 82 labelled images}
    \label{fig:spectra}
\end{figure}
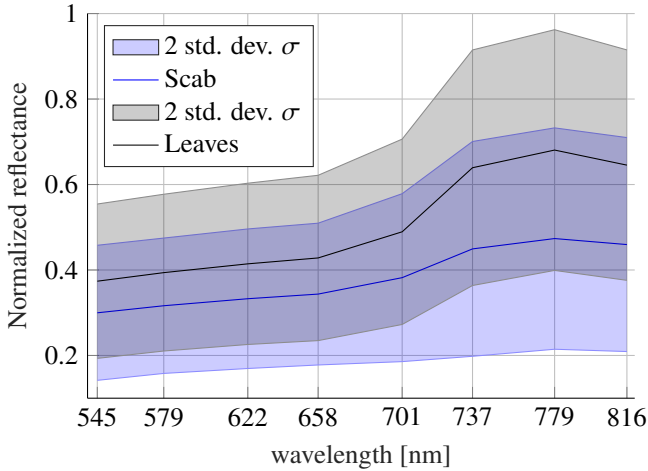

Figure \ref{fig:spectra} shows the spectral data for two selected areas of the images, the scab and the healthy leaves. These graphs were created from 170\,000\,pixels for each class from randomly sampled 82 labelled images throughout the entire data set. These images were expertly annotated and mean values plotted. Two shaded areas represent borders of two standard deviations, and it is apparent that there is overlap in the spectral signatures of healthy and unhealthy leaves. Because of this, a more elaborate algorithm such as multilayer perceptron was required to separate the classes in the multidimensional image.

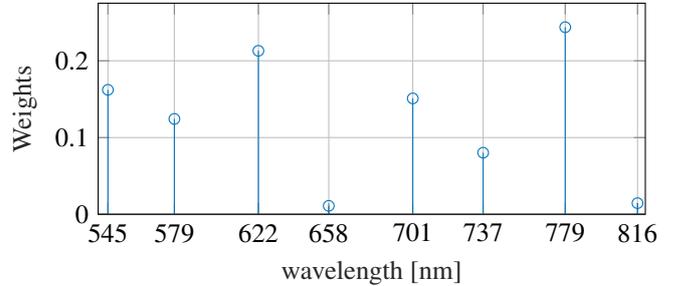
\begin{figure}[htbp]
    \centering
%
%
\definecolor{mycolor1}{rgb}{0.00000,0.44700,0.74100}%
\begin{tikzpicture}

\begin{axis}[%
width=7.2cm,
height=1.102in,
at={(0.395in,0.304in)},
scale only axis,
xmin=540,
xmax=820,
xtick={545, 579, 622, 658, 701, 737, 779, 816},
xlabel style={font=\color{white!15!black}},
xlabel={wavelength [nm]},
ymin=0,
ymax=0.275,
ylabel style={font=\color{white!15!black}},
ylabel={Weights},
axis background/.style={fill=white},
title style={font=\bfseries},
xmajorgrids,
ymajorgrids
]
\addplot[ycomb, color=mycolor1, mark=o, mark options={solid, mycolor1}, forget plot] table[row sep=crcr] {%
545	0.162178593896405\\
579	0.124243329432665\\
622	0.213010504373986\\
658	0.0110633743701111\\
701	0.150999428689756\\
737	0.0802314780019744\\
779	0.24373327527784\\
816	0.0145400159572631\\
};
\addplot[forget plot, color=white!15!black] table[row sep=crcr] {%
540	0\\
820	0\\
};
\end{axis}

\begin{axis}[%
width=3.937in,
height=1.575in,
at={(0in,0in)},
scale only axis,
xmin=0,
xmax=1,
ymin=0,
ymax=1,
axis line style={draw=none},
ticks=none,
axis x line*=bottom,
axis y line*=left
]
\end{axis}
\end{tikzpicture}%
    \caption{LDA discriminative power from 45\,000 randomly sampled scab and heatlhy pixels of labelled MS images. Higher the weight is higer the discriminative power.}
    \label{fig:LDA}
\end{figure}

A subset of 45\,000 pixels from previously plotted data for each class was used to train the LDA. LDA was used to find the most discriminating channels using leaf pixels against scab pixels. Using the LDA weight vector areas of the spectrum that contribute the most to discrimination, this weighting vector can be seen in the Figure \ref{fig:LDA}. The confusion matrix for LDA is given in Table \ref{tab:LDA-CM}. An ordered vector of wavelengths from most to least discriminating was created to be used in the generation of false-colour images.

The order of wavelengths from most to least discriminating was 779\,nm, 622\,nm, 545\,nm, 701\,nm, 579\,nm, 737\,nm, 816\,nm, 658\,nm.

\begin{table}[htbp]
\caption{LDA confusion matrix for MS image pixels}
\begin{center}
    \begin{tabular}{@{}lrrr@{}}
    \toprule
    True labels     & Leaf      & Scab      & Totals \\ \midrule
    Leaf            & 0.820     & 0.180     & 1      \\
    Scab            & 0.127     & 0.873     & 1      \\ \bottomrule
    \end{tabular}
\end{center}
    \label{tab:LDA-CM}
\end{table}

\begin{figure}[htbp]
    \centering
    \includegraphics[width=0.48\textwidth]{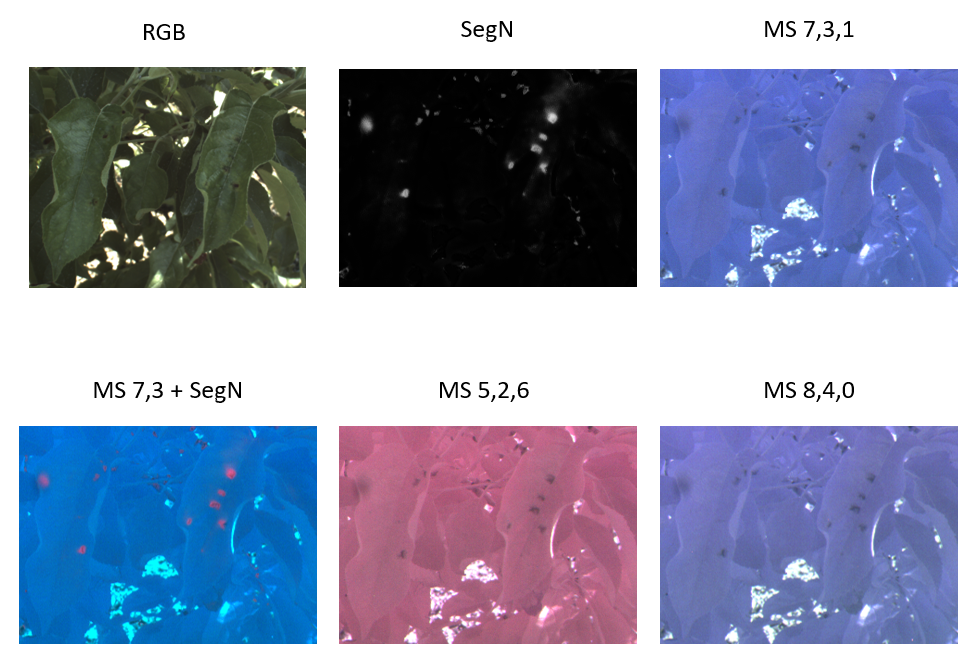}
    \caption{Examples of each of the six image sets used in this experiment. The top left is a traditional RGB images, the top center is a monochrome image of the Segmentation Network Results. The top right is am false colour image composed of the multispectral bands 7,3, and 1. The Bottom left is a false colour composed of multispectral bands 7 and 3 and the results of the segmentation network. the bottom center is a representation of a 6 band image composed of multispectral bands 7,3,1,5,2 and 6. and the bottom right is a representation of a 9 band image comprising of all multispectral bands.}
    \label{fig:falseCombinations}
\end{figure}

Five combinations of false-colour images were created from spectral data and the segmentation net result. The preview of them is shown in figure \ref{fig:falseCombinations}.

\subsection{Deep learning results}
\begin{figure}[htbp]
    \centering
    \includegraphics[width=0.48\textwidth]{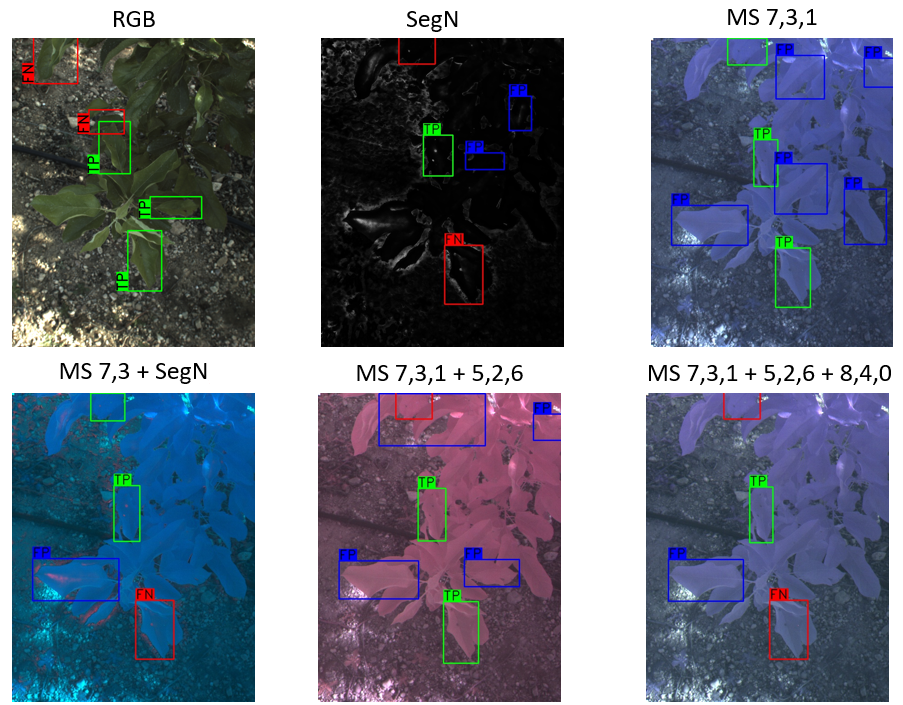}
    \caption{The results of the YOLO Network on 6 different image sets created from the same camera perspective. The top left is a traditional RGB image, the top center is a monochrome image of the segmentation network. The top right is a false colour image composed of the multispectral bands 7,3, and 1. The Bottom left is a false colour composed of multispectral bands 7 and 3 and the results of the segmentation network. the bottom center is a representation of a 6 band image composed of multispectral bands 7,3,1,5,2 and 6. and the bottom right is a representation of a 9 band image comprising of all multispectral bands. The Bounding boxes (TP - true positive, FP - false positive, FN - false negative) represent scab detection in each of the six example image sets.}
    \label{fig:montage}
\end{figure}

Table \ref{tab:table_det_results} and Figure \ref{fig:montage} show the results of the deep learning-based object detection. The best object detection performance was achieved with the RGB images (image set 1). The F1 score when using RGB images was 0.16 to 0.21 higher than when using multispectral image sets. Both mAP scores were also significantly better when using RGB images (Table \ref{tab:table_det_results}). This result was not expected given the better colour contrast of the multispectral images compared to the RGB images (Figure \ref{fig:falseCombinations}).

\begin{table}[hbt!]
\caption{Object detection results for the six image sets (refer to Table \ref{tab:ListIma}). P, R, F1 and mAP are abbreviations of precision, recall, F1-score and mean average precision, respectively. mAP@0.5 is the mAP value when using an IoU threshold of 0.5. mAP@..0.95 is the average mAP value when ranging the IoU threshold between 0.5 and 0.95 in steps of 0.05.}
\begin{center}
 \begin{tabular}{l c c c c c} 
 \hline
  & & & & mAP & mAP\\
 Image set & P & R & F1 & @0.5 & @..0.95\\ 
 \hline
 RGB & \textbf{0.80} & \textbf{0.63} & \textbf{0.70} & \textbf{0.73} & \textbf{0.56}\\
 SegN & 0.49 & 0.49 & 0.49 & 0.47 & 0.24\\
 MS7,3,1 & 0.48 & 0.56 & 0.51 & 0.53 & 0.35\\
 MS7,3+SegN & 0.52 & 0.46 & 0.49 & 0.43 & 0.29\\
 MS7,3,1+5,2,6 & 0.72 & 0.43 & 0.54 & 0.52 & 0.37\\
 \shortstack[l]{MS7,3,1+5,2,6\\+8,4,0} & 0.53 & 0.51 & 0.52 & 0.47 & 0.30\\
 \hline
 \end{tabular}
 \end{center}
 \label{tab:table_det_results}
\end{table}

\subsection{Discussion}
Despite the promising results of deep learning using RGB images, the detection of apple scab in apple trees using multispectral imaging proved to be a difficult task. While the deep learning model trained on the spectral images was able to detect some general patterns, the model was unable to detect disease with the same accuracy as the YOLO trained on the RGB images. This is an interesting result, as the spectral images showed a much higher visual contrast between the apple scab lesions and healthy leaves compared to the RGB images (Figure \ref{fig:falseCombinations}). This may be due to the use of pretrained networks in this study. Pretrained networks use RGB image sets, such as MS COCO, to initialise the weights of the network before training. If the false-colour images were not similar enough to these images, then that could explain the outstanding performance of RGB images over false-colour.
Furthermore, the results may be affected by the outdoor environment in which apple trees are grown. False positives in the images were shown to be caused by shadows on leaves and other diseases such as leaf miners. The high-light environment of the open field made it difficult to collect a balanced spectrum from the multispectral camera as the infrared channel and the visible channels needed to be constantly balanced so that they did not overexpose in the images.
 These results suggest that more research is needed to deploy deep learning networks to spectral datasets. Further research is needed to determine whether more detailed data may be needed to improve the accuracy of deep learning models or if more finely tuned models are needed to handle spectral image data.

\section{Conclusion}
In this experiment, the efficacy of deep learning detection of apple scab disease was compared on multispectral and traditional RGB images. It was found that using multispctral imaging can be used to increase the contrast drastically between healthy and infected tissue; however, counter-intuitively, this did not lead to better results from the neural network detector.

Several hypotheses for this were considered, including the small size of the dataset, or the effects of transfering learning on multispectral images that had been based on RGB colour images, specifically the COCO dataset.

The higher contrast in multispectral images still suggests that they may be used in the future with neural networks with further groundwork. We will continue to explore this topic in future research and encourage other researchers to do so as well.

\section*{CRediT authorship contribution statement}
 Robert Rou\v{s}: Software, Investigation, Methodology, Writing - Original Draft, Visualization; Joseph Peller: Methodology, Software, Writing - Original Draft \& Editing; Gerrit Polder: Conceptualisation, Supervision, Project administration, Writing – Review \& Editing; Selwin Hageraats: Software, Writing - Original Draft; Thijs Ruigrok: Software, Writing - Review \& Editing; Pieter M. Blok: Software, Data Curation, Writing - Review \& Editing.

\section*{Acknowledgement}\label{sec:acknowledgement}
This paper is supported by European Union’s Horizon 2020 research and innovation programme under grant agreement No 773718, project OPTIMA (Optimised Pest Integrated Management to precisely detect and control plant diseases in perennial crops and open-field vegetables). We would like to acknowledge the following people from the Universitat Politècnica de Catalunya (UPC):  Emilio Gil (UPC), Paula Ortega (UPC), Fran Garcia (UPC).



\bibliographystyle{elsarticle-harv} 
\bibliography{bibliography}





\end{document}